\documentclass[a4paper,10pt]{article}
\usepackage{times}
\usepackage{graphicx}\graphicspath{{.}} 
\usepackage{epstopdf}
\usepackage{algorithm}
\usepackage{algorithmic}
\usepackage{multirow} 
\usepackage{multicol}
\usepackage[cmex10]{amsmath}
\usepackage{array}
\usepackage{mdwmath}
\usepackage{mdwtab}
\usepackage{rotating}
\usepackage{caption}
\usepackage{subfig}
\usepackage{amssymb}
\usepackage[inline]{enumitem}\setlist[enumerate]*{label=\arabic*.}
\newcommand{\il}[1]{\begin{enumerate*}[label=(\roman*)]#1\end{enumerate*}}
\usepackage{verbatim}

\newcommand{\fref}[1]{Fig.~\ref{#1}}       
\newcommand{\sref}[1]{\S\ref{#1}}          
\newcommand{\tref}[1]{\tablename~\ref{#1}} 
\newcommand{\eref}[1]{(\ref{#1})}          

\newcommand{\eg}[1]{\textit{e.g.,}~#1} %
\newcommand{\ie}[1]{\textit{i.e.,}~#1} %

\usepackage{textcomp} 

\newcommand{\estimated} [1]{\tilde{#1}}        
\newcommand{\normalised}[1]{\hat{#1}}          
\newcommand{\pinv}      [1]{#1^\dagger}        

\newcommand{\R}       {\mathbb{R}}             
\newcommand{\T}       {\top}                   
\newcommand{\I}       {\mathbf{I}}             
\newcommand{\bO}      {\boldsymbol{0}}         
\newcommand{\U}       {\mathcal{U}}            

\newcommand{\dimX}    {\mathcal{P}}            
\newcommand{\dimU}    {\mathcal{\MakeUppercase{\nu}}} 
\newcommand{\dimR}    {\mathcal{S}}            
\newcommand{\dimB}    {\mathcal{\MakeUppercase{\nb}}} 
\newcommand{\dimPhi}  {\mathcal{\MakeUppercase{\nphi}}} 

\newcommand  {\nd}      {n}                    
\renewcommand{\nu}      {q}                    
\newcommand  {\nb}      {s}                    
\newcommand  {\nphi}    {m}                    

\newcommand  {\Nd}      {\mathcal{\MakeUppercase{\nd}}}  

\newcommand  {\bx}    {\mathbf{x}}             
\newcommand  {\bu}    {\mathbf{u}}             
\newcommand  {\bpi}   {\boldsymbol{\pi}}       
\newcommand  {\btheta}{\boldsymbol{\theta}}    

\newcommand  {\bq}      {\mathbf{q}}           
\newcommand  {\bqdot}   {\dot{\bq}}            
\renewcommand{\r}       {r}                    
\newcommand  {\br}      {\mathbf{\r}}          
\newcommand  {\bJ}      {\mathbf{J}}           
\newcommand  {\bA}      {\mathbf{A}}           
\newcommand  {\bb}      {\mathbf{b}}           
\newcommand  {\bN}      {\mathbf{N}}           
\newcommand  {\buts}    {\bu^\mathrm{ts}}      
\newcommand  {\buns}    {\bu^\mathrm{ns}}      

\newcommand  {\bxn}     {\bx_\nd}              
\newcommand  {\bun}     {\bu_\nd}              
\newcommand  {\butsn}   {\buts_{\nd}}          
\newcommand  {\bunsn}   {\buns_{\nd}}          
\newcommand  {\bAn}     {\bA_\nd}              
\newcommand  {\bbn}     {\bb_\nd}              
\newcommand  {\bNn}     {\bN_\nd}              
\newcommand  {\bpin}    {\bpi_\nd}             
\newcommand  {\ebutsn}  {\ebuts_{\nd}}         
\newcommand  {\ebunsn}  {\ebuns_{\nd}}         
\newcommand  {\ebPn}    {\ebP_\nd}             
\newcommand  {\ebAn}    {\ebA_\nd}             
\newcommand  {\ebNn}    {\ebN_\nd}             

\newcommand  {\bP}      {\mathbf{P}}           

\newcommand  {\ebu}     {\estimated{\bu}}      
\newcommand  {\ebuts}   {\ebu^{\mathrm{ts}}}   
\newcommand  {\ebuns}   {\ebu^{\mathrm{ns}}}   
\newcommand  {\ebA}     {\estimated{\bA}}      
\newcommand  {\ebN}     {\estimated{\bN}}      
\newcommand  {\ebP}     {\estimated{\bP}}      

\newcommand  {\bL}      {\mathbf{L}}           

\newcommand{\ba}      {\boldsymbol{\alpha}}    %
\newcommand{\nba}     {\normalised{\ba}}    %
\newcommand{\na}     {\normalised{a}}    %

\DeclareMathOperator{\state}{\mathbf x}
\DeclareMathOperator{\action}{\mathbf u}
\DeclareMathOperator{\buTs}{{\mathbf u}^{ts} }
\DeclareMathOperator{\buNs}{{\mathbf u}^{ns} }

\DeclareMathOperator{\NullspacePolicy}{{\boldsymbol{\pi}}}
\DeclareMathOperator{\Basis}{{\boldsymbol{\beta}}}
\DeclareMathOperator{\TaskspacePolicy}{{\mathbf b}}
\DeclareMathOperator{\Degree}{^{\text{\textdegree}}}

\DeclareMathOperator{\Real}{{\mathbb R}}

\DeclareMathOperator{\LearntU}{\tilde{\action}}

\DeclareMathOperator{\LearntUts}{\LearntU^{ts} }
\DeclareMathOperator{\LearntUns}{\LearntU^{ns}}
\DeclareMathOperator{\LearntN}{\tilde{\mathbf N}}

\DeclareMathOperator{\Utsn}{\tilde{\action}^{ts}_n}
\DeclareMathOperator{\Unsn}{\tilde{\action}^{ns}_n}

\DeclareMathOperator{\WeightsMatrix}{\mathbf W}
\DeclareMathOperator{\WeightsVector}{\mathbf w}
\DeclareMathOperator{\Jacobian}{{\mathbf J} }

\DeclareMathOperator{\Identity}{{\mathbf I} }
\DeclareMathOperator{\Transpose}{{^{\top}} }

\newcommand{\bLambda} {\boldsymbol{\Lambda}}       %

\usepackage{color}
\usepackage[normalem]{ulem}                                        

\definecolor{orange}{RGB}{255,128,0}

\newcommand{\vectornorm}[1]{||#1||}

\makeatletter\newcommand{\mylabel}[2]{\def\@currentlabel{#2}\label{#1}}\makeatother

\title{\LARGE\bf Learning Null Space Projections in Operational Space Formulation}

\author{%
Hsiu-Chin Lin and Matthew~Howard
\thanks{%
H. Lin({\tt\small H.Lin@bham.ac.uk}) is at the School of Computer Science, University of Birmingham, UK. M. Howard ({\tt\small matthew.j.howard@kcl.ac.uk}) is at the Dept. of Informatics, Kings College London, UK.
}%
}%

\begin{document}
\setlength{\floatsep}{5pt}          
\setlength{\textfloatsep}{7pt}      
\maketitle
\thispagestyle{empty}
\pagestyle{empty}
\begin{abstract}
In recent years, a number of tools have become available that recover the underlying control policy from constrained movements. However, few have explicitly considered learning the constraints of the motion and ways to cope with unknown environment. In this paper, we consider learning the null space projection matrix of a kinematically constrained system in the absence of any prior knowledge either on the underlying policy, the geometry, or dimensionality of the constraints. Our evaluations have demonstrated the effectiveness of the proposed approach on problems of differing dimensionality, and with different degrees of non-linearity.
\end{abstract}

\section{Introduction}      \label{introduction}      
\begin{figure}[t!]
  \centering
  \includegraphics[height=2cm]{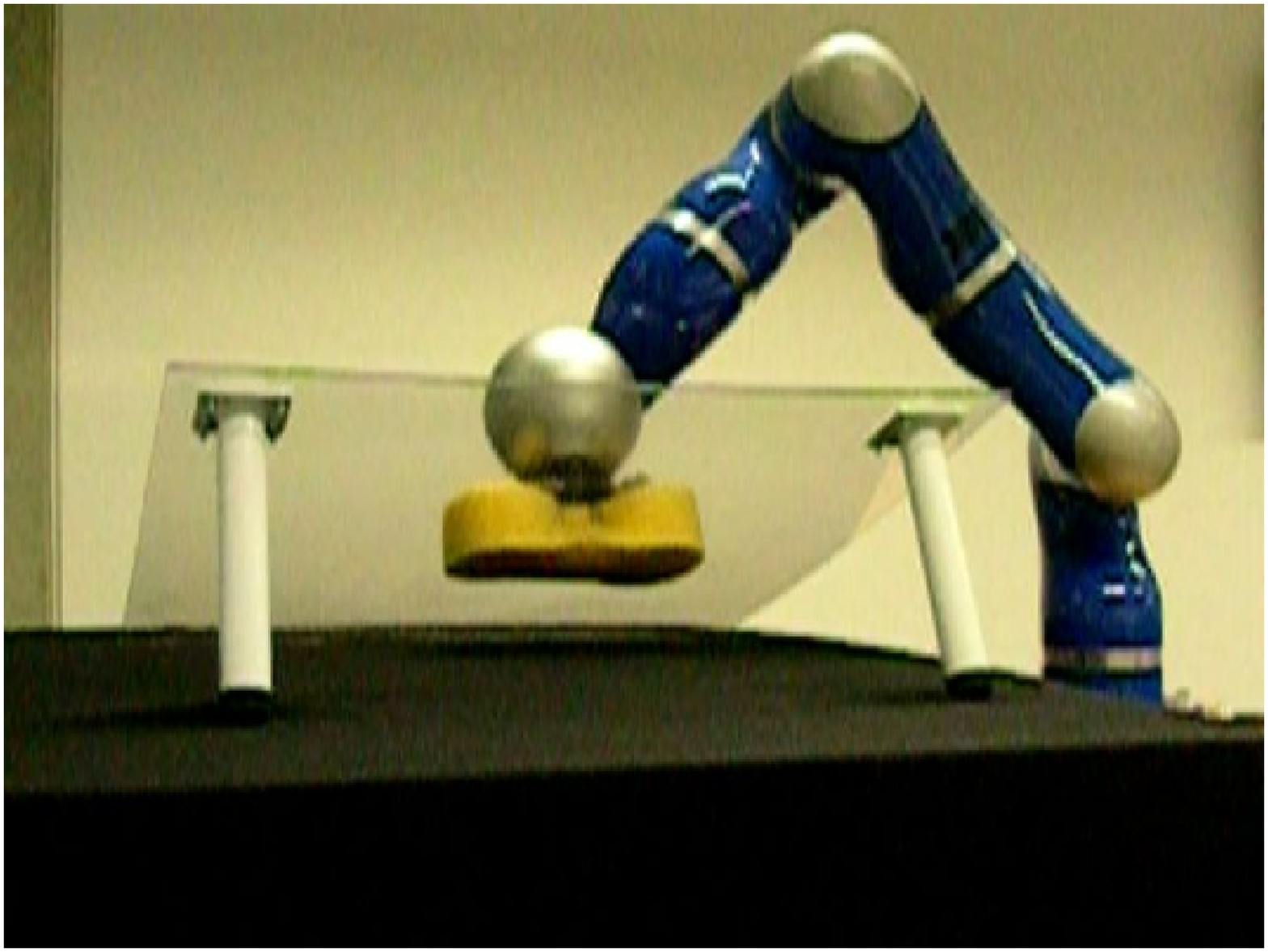} 
  \includegraphics[height=2cm]{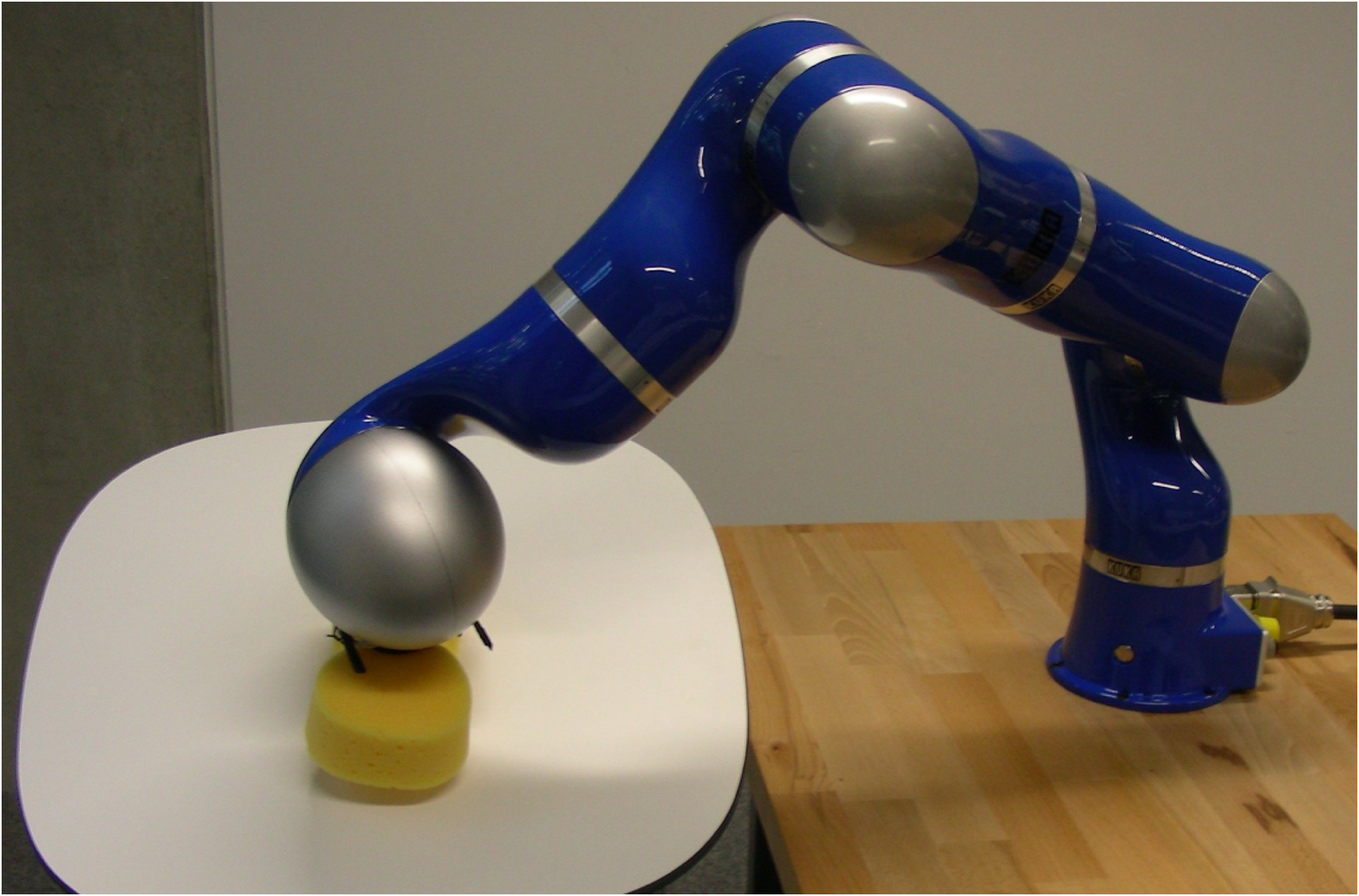} 
  \includegraphics[height=2cm]{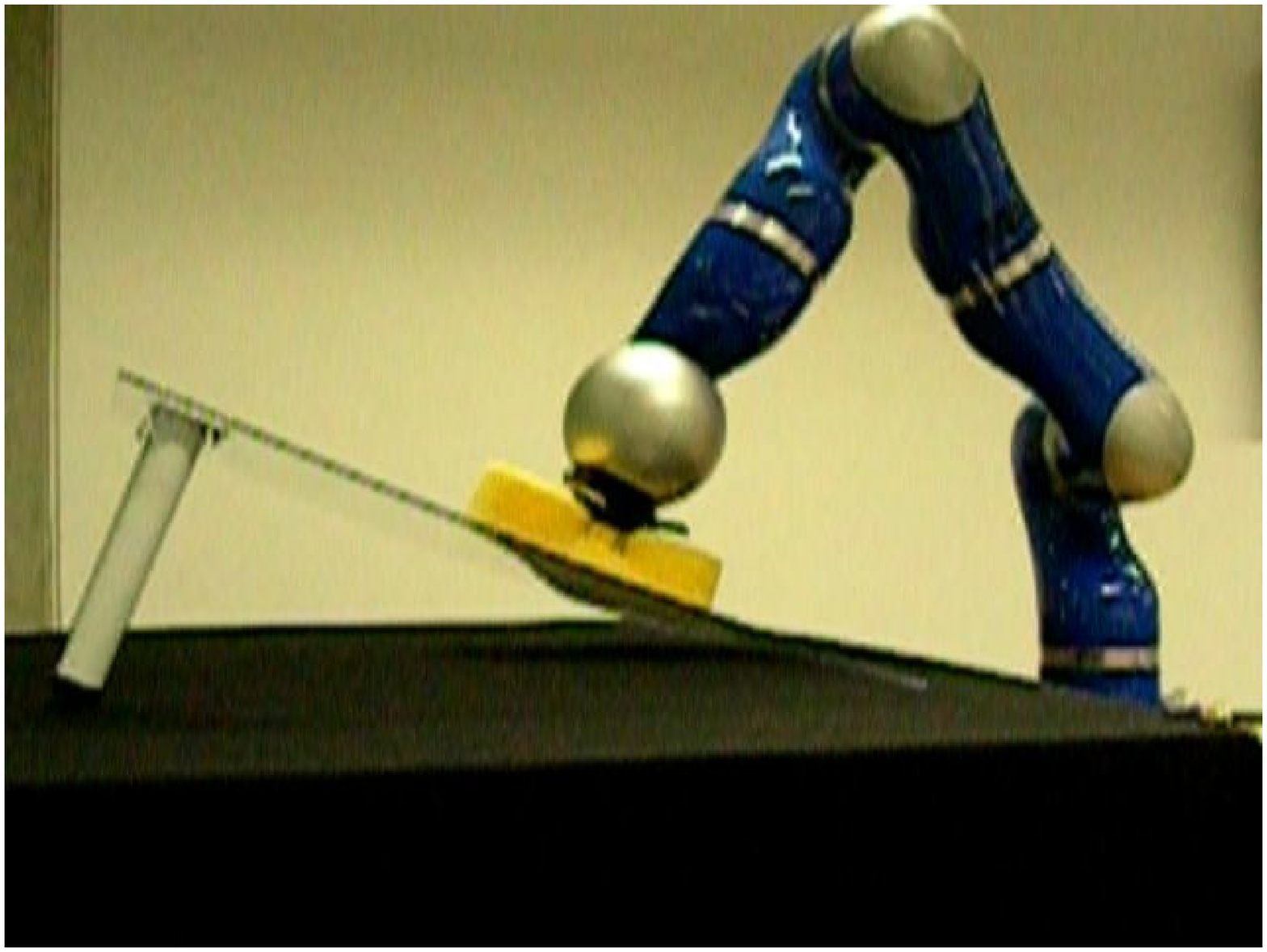} \\ \vspace{-1mm}
  \caption{Examples of wiping on different tables~\cite{Howard2008}. The behaviour (wiping) is subject to various constraint imposed by the environment where the behaviour is performed (surfaces).}
  \label{fig:intro} 
\end{figure}

\noindent 
Many everyday human skills can be considered in terms of performing some task subject to a set of self-imposed or environmental constraints. For example, when pouring water from a bottle, self-imposed constraints apply to the
position and the orientation of the hand so that the water falls within the glass. When wiping a table (Fig.~\ref{fig:intro}), the surface of the table acts as an environmental constraint that restricts the hand movements when in contact with the surface. 

A promising way to provide robots with skills is to take examples of human demonstrations and attempt to learn a control policy that somehow capture the behaviours~\cite{Billard.2007,Craig.1987.IJRR,Kawato.1990.ANC}. One common approach is to take the operational space formulation~\cite{1987.IJRR.Khatib}. For example, given constraint in the end-effector space, produce a set of joint-space movements that can satisfy the constraints. Behaviour may be subject to various constraints that are usually non-linear in actuator space~\cite{Gienger.2005.Humanoids,Khatib.2004.IJRR}. For example, maintaining balance of the robot (higher priority) while accomplishing an end-effector task (lower priority)~\cite{Sentis.2005.IJHR} or avoiding obstacles~\cite{Stilman.2008.IJRR}.
 
In recent years, a number of new tools have become available for recovering the underlying policy from constraint data~\cite{Howard.2009.AR,Towell.2010.IROS}; however, few have explicitly considered learning the {\em constraints} and coping with unknown environment. Previous work related to the estimation of constraint takes force measurements from the end-effector to calculate the plane normal to the constraint~\cite{Blauer.1987,Yoshikawa.1987}. Nevertheless, these are limited to problems of a robot manipulator acting on a smooth surface in a three-dimensional space, and rely on force sensors, which are normally expensive to obtain. 


In this paper, we propose a method for directly learning the kinematic constraints present in movement observations, as represented by the null space projection matrix of a kinematically constrained system. The proposed approach requires no prior information about either the dimensionality of the constraints, nor does it require information about the policy underlying the observed movement.

\section{Problem Definition}\label{problem_definition}\noindent 
Based on the principles of analytical dynamics~\cite{Udwadia.2007}, we consider
that the underlying policy is subject to a set of $\dimB$-dimensional
($\dimB\le\dimU$) constraints

\begin{equation}
  \bA(\bx)\bu(\bx)=\TaskspacePolicy(\bx)
  \label{equ:problem-with-task} 
\end{equation}

\noindent where $\bx\in\Real^\dimX$ represents state, $\bu\in\Real^\dimU$ represents
the action, and $\TaskspacePolicy\neq 0$ is the \emph{task-space
policy} describing the underlying task to be accomplished. The \emph{constraint
matrix} $\bA(\state)\in\Real^{\dimB\times\dimU}$ is a matrix describing the
constraints, which projects the task-space policy onto the relevant part of the
control space. Inverting \eref{equ:problem-with-task}, results in the relation 
\begin{equation}
  \bu(\bx)=\pinv{\bA}(\bx)\bb(\bx)+\bN(\bx)\bpi(\bx)
  \label{equ:solution-with-task}
\end{equation}
\noindent where $\pinv{\bA}$ is the Moore-Penrose pseudo-inverse of $\bA$,

\begin{equation}
  \bN(\bx):=(\Identity-\pinv{\bA}(\bx)\bA(\bx))\in\Real^{\dimU\times\dimU} 
  \label{N}
\end{equation}
\noindent is the projection matrix, and $\Identity\in\Real^{\dimU\times\dimU}$ is the
identity matrix. The projection matrix $\bN$ projects the null space policy $\bpi$
onto the null space of $\bA$, which in general, has non-linear dependence on
both time and state. Note that, as $\bN$ projects $\bpi$ onto the null space of $\bA$, the two terms
\begin{equation}
	\buts(\bx):=\pinv{\bA}(\bx)\bb(\bx) \text{ and } \buns(\bx):=\bN(\bx)\bpi(\bx) 
	\label{e:buns}
\end{equation}
are orthogonal to each other. We term them as the \emph{task space component} and \emph{null space component}, respectively. 

It would be useful to know the decomposition of $\bA$, $\TaskspacePolicy$, $\bN$, and $\NullspacePolicy$; however, the
true quantities of those variables are unavailable by assumption. In human demonstrations, it is normally not clear which dimensions are restricted as part of the task constraints. For example, when picking up a glass of water, a reasonable assumption is that the orientation of the hand is controlled such that the glass is kept up right. However, it is less clear whether the hand position is part of the task (result of $\TaskspacePolicy$) or a comfortable position is chosen as part of the redundancy resolution (result of $\NullspacePolicy$). 

Several studies have been devoted to learning $\buNs$ (or, equivalently, $\buTs$) and $\bpi$~\cite{Towell.2010.IROS}, but so far, few have explicitly considered estimating $\bA$ or $\bN$. However, the ability to estimate $\bA$ or $\bN$ is favourable for adaptation. When dealing with an unseen constraint (i.e., a different $\bA$), it might be time-consuming to re-learn a policy $\bpi$ and require lots of additional human demonstrations. For example, adapting the wiping behaviour onto a different table (see Fig. 1) requires prior knowledge of the surface itself. If the constraint (the surface) can be estimated, we can adapt the previously learnt wiping behaviour onto the new constraint.


In~\cite{Lin.2014.Robotica}, it was first demonstrated that $\bN$ could be learnt purely from data for the case $\dimU=2$. Subsequently, \cite{Lin.2015.ICRA} showed that $\bN$ can be estimated for problems where $\bA\in\R^{\dimB\times\dimU}$ and $1<\dimB<\dimU$, for the special case of $\bA\bu=0$. By extending that work, here we propose a method to estimate $\bN$ for the generic case described by \eref{equ:problem-with-task} (with $\bb\neq0$), the first time this has been shown for problems in the full operational space formulation \eref{equ:problem-with-task}-\eref{equ:solution-with-task}.       
\section{Method}            \label{method}            
\noindent The proposed method works on data given as $\Nd$ pairs of observed
states $\bxn$ and observed actions $\bun$. It is assumed that 
\il{
\item the observations follow the formulation in \eref{equ:solution-with-task}, 
\item the task space policy $\bb$ varies across observations,
\item $\bu$ are generated using the same null space policy $\bpi$, 
\item neither $\bA$, $\bb$ nor $\bN$ are explicitly known for any given observation.
} We define the shorthand notation $\pinv{\bAn}:=\pinv{\bA}(\bxn)$, $\bbn:=\bb(\bxn)$, $\bNn:=\bN(\bxn)$ and $\bpin:=\bpi(\bxn)$).

\subsection{Learning Null Space Component}
\label{ss:learning_null_space_component}
\noindent
The first step is to extract an estimate of the null space component \eref{e:buns} from the raw observations. From~\cite{Towell.2010.IROS}, an estimate $\ebuns(\bx)$ is sought which minimises
\begin{equation} 
	E[\ebuns]=\sum_{\nd=1}^{\Nd}\vectornorm{\ebPn\bun-\ebunsn}^2  
  \label{equ:learn-uns} 
\end{equation}
\noindent 
where $\ebunsn:=\ebuns(\bxn)$ and $\ebPn:=\ebunsn\ebunsn\,^\T/\vectornorm{\ebunsn}^2$. This exploits the identity $\bP\bu=\bP(\buts+\buns)=\buns$, by seeking a estimate consistent with this, see \cite{Towell.2010.IROS} for details.

\subsection{Learning Null Space Projections}
\noindent 
Having an estimate of the null space term $\ebuns$, and knowing that the data follows the relationship \eref{equ:solution-with-task}, allows several properties of \eref{equ:solution-with-task} to be used to form the estimate of $\bA$. 

\begin{figure}[t!]
    \centering
    \subfloat[] {\label{fig:obj-data}	\includegraphics[width=4.25cm]{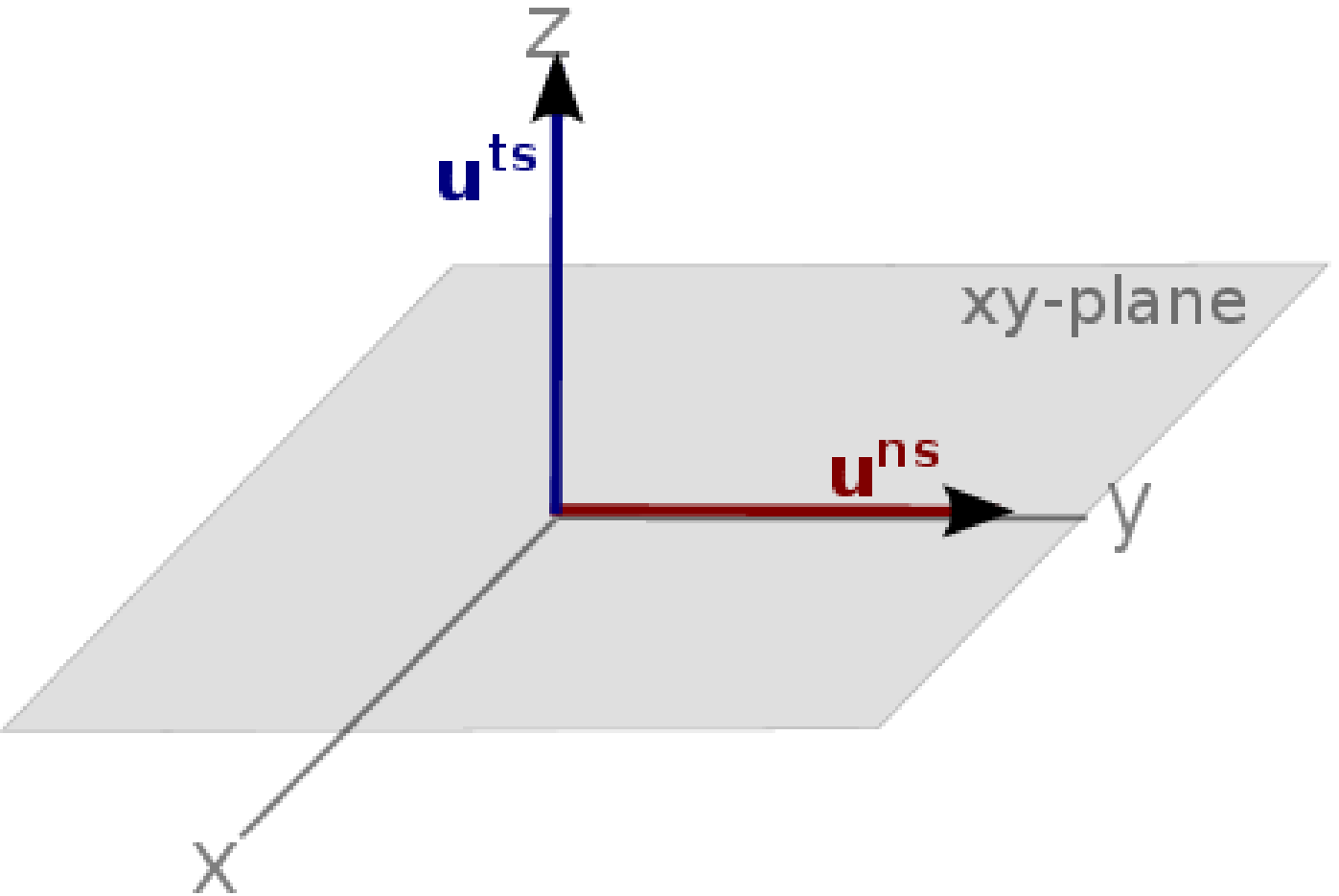}} 
    \subfloat[] {\label{fig:obj-uns}	\includegraphics[width=4.25cm]{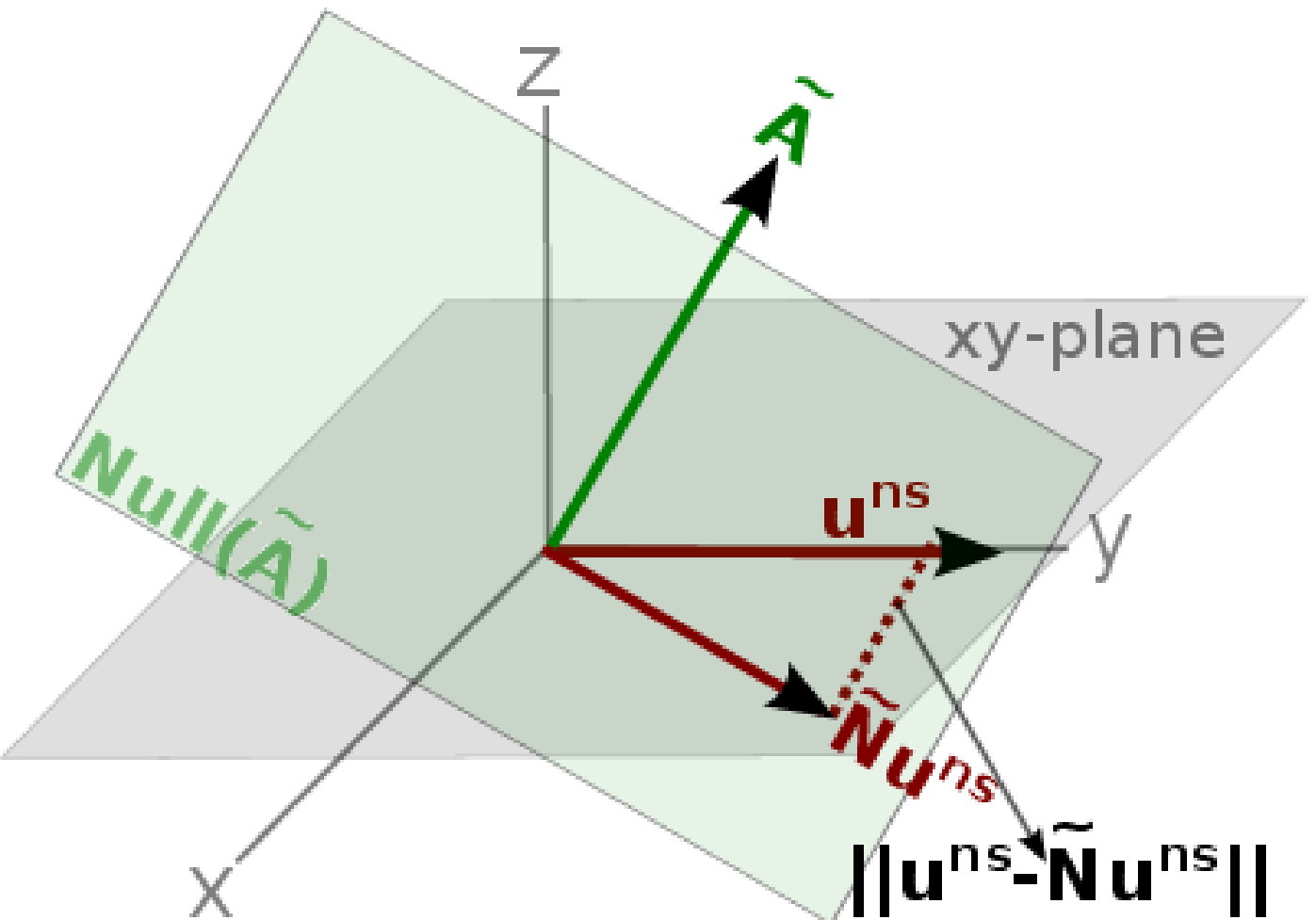}}  \\ \vspace{-3mm}
    \subfloat[] {\label{fig:obj-uts}	\includegraphics[width=4.25cm]{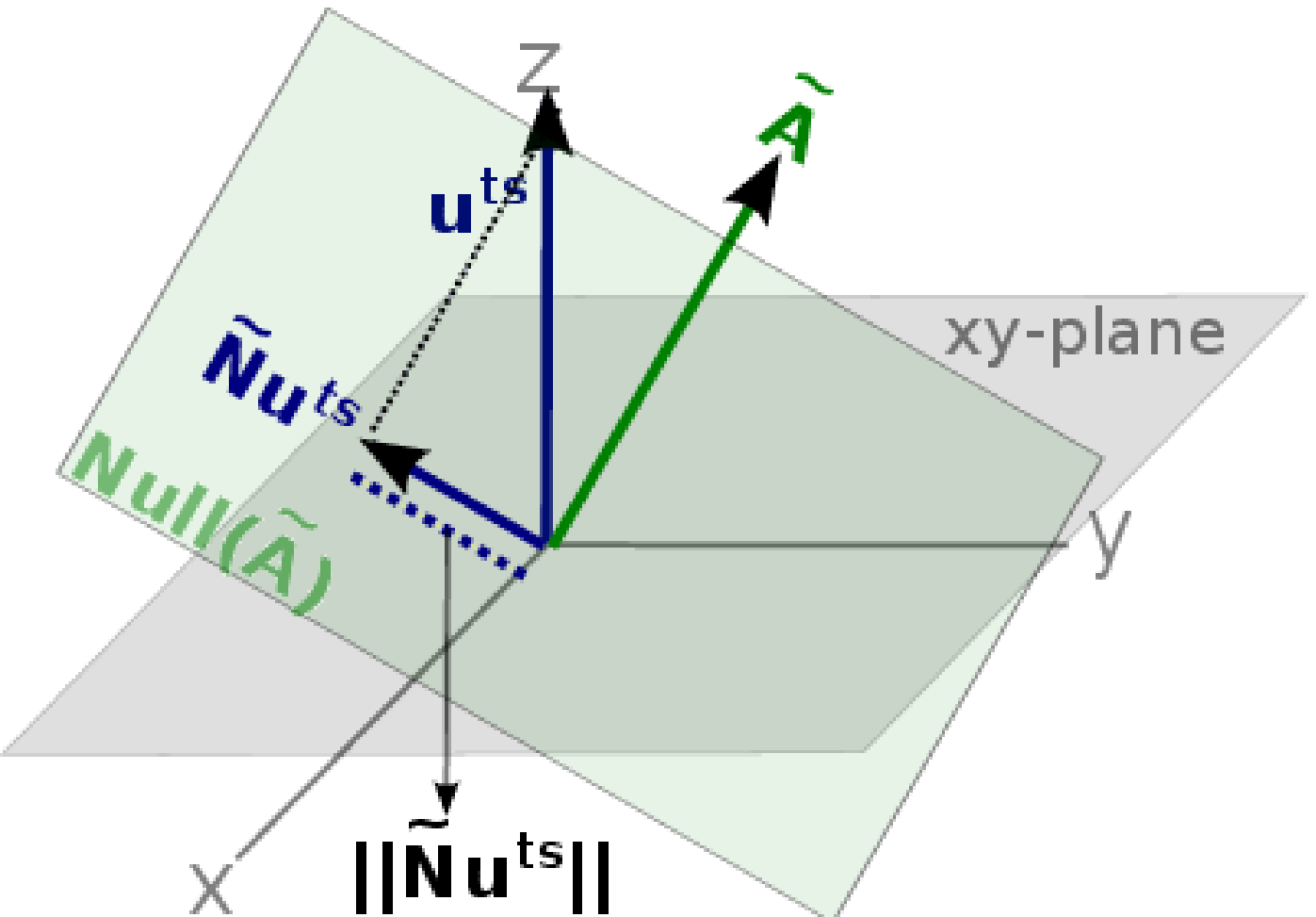}}
    \subfloat[] {\label{fig:obj-uts-uns}\includegraphics[width=4.25cm]{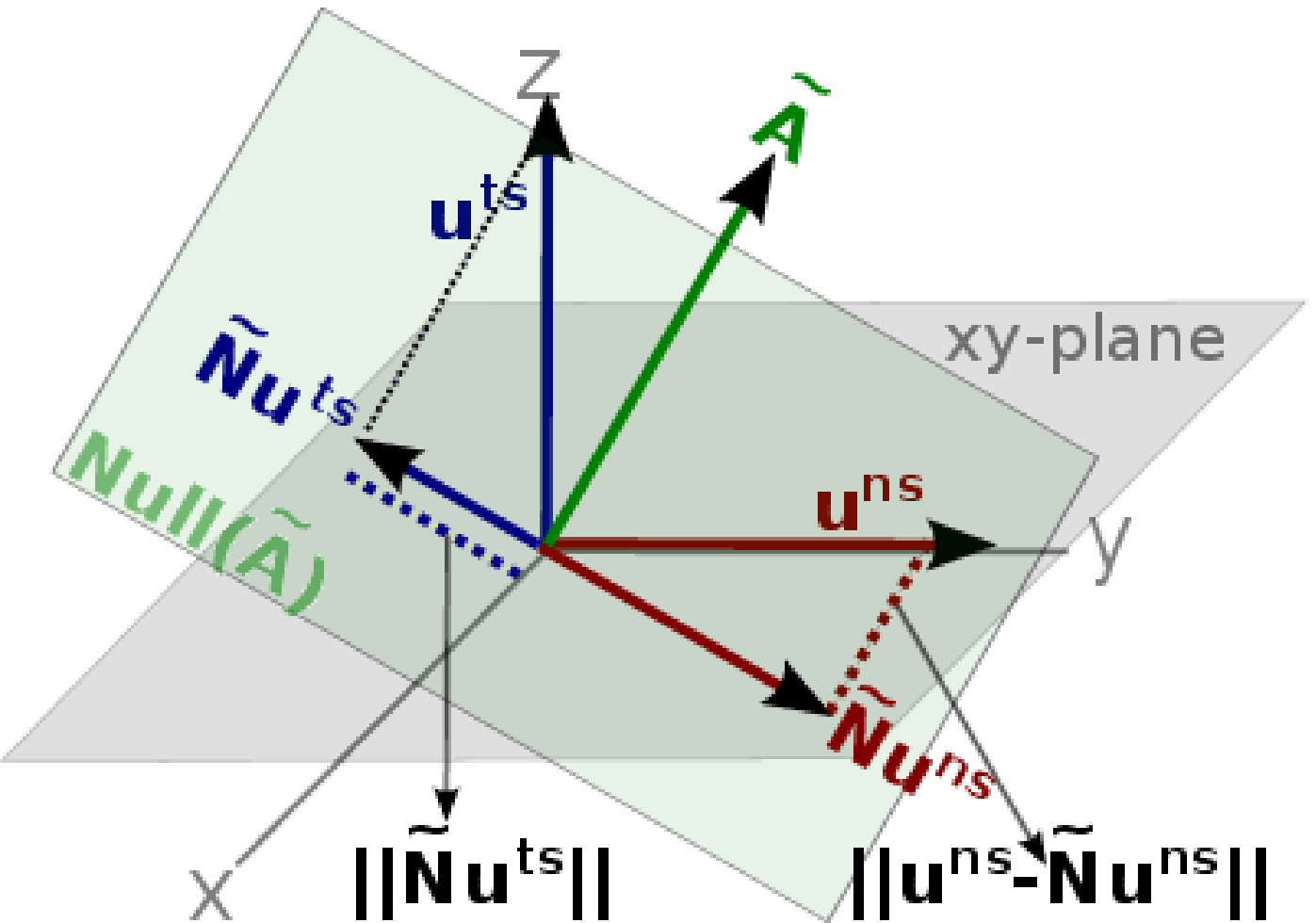}}  \\   \vspace{-2mm} 
	\caption{Visualisation of the objective functions. (a) Example data point with
		task and null space components plotted. (b) The objective
		function~\eref{equ:learn-n-uns} minimises
		$\vectornorm{\buns-\ebN\buns}$ (red dashed line). (c) Alternately,
		\eref{equ:learn-n-uts} minimises
		$\vectornorm{\ebN\buts}$ (blue dashed line).
		(d) It is proposed to use~(\ref{equ:learn-n-with-tasks}), which
		minimises the sum of these distances.}
    \label{fig:obj}
\end{figure}

Firstly, since $\buns$ is the projection of $\bpi$ onto the image space of $\bN$ (see \eref{e:buns}), and by the indempotence of $\bN$,
\begin{equation}
	\bN\buns=\buns
	\label{e:Nuns=uns}
\end{equation}
must hold~\cite{Lin.2015.ICRA}. This means that an estimate $\ebN$ may be furnished by optimising \eref{e:Nuns=uns}, \ie minimising
\begin{equation}
  E[\ebN]=\sum_{\nd=1}^\Nd\vectornorm{\ebunsn-\ebNn\ebunsn}^2
  \label{equ:learn-n-uns}
\end{equation}
where $\ebNn:=\ebN(\bxn)$. \fref{fig:obj-data}-\ref{fig:obj-uns} shows a visualisation of this objective function. In \fref{fig:obj-data}, an example data point is plotted where $\bA$ is a vector parallel to the $z$-axis, its null space is the $xy$-plane, the null space component $\buns$ is parallel to the $y$-axis, and the task space component $\buts$ is parallel to the $z$-axis. The objective function~\eref{equ:learn-n-uns} aims at minimising the distance between $\buns$ and its projection onto the null space of $\bA$ (green plane), illustrated as the red dashed line. 

Second, note that, by definition, $\buts$ and $\buns$ are orthogonal (\ie $\buts\,^\T\buns=0$) and so the true projection matrix must also satisfy $\bN\buts=\bO$. Using this insight, another alternative is to seek an estimate $\ebN$ that minimises
\begin{equation}
   E[\ebN]=\sum_{\nd=1}^\Nd \vectornorm{\ebN\ebutsn}^2
  \label{equ:learn-n-uts}
\end{equation} 
where $\ebutsn:=\ebuts(\bxn)$. A visualisation of~(\ref{equ:learn-n-uts}) for the example data point is shown in \fref{fig:obj-uts}, where now the blue dashed line indicates the distance minimised.

Since both~(\ref{equ:learn-n-uns}) and~(\ref{equ:learn-n-uts}) contain information about the projection matrix, the third alternative, proposed here is to minimise the sum of two, namely, 
\begin{equation}
  E[\ebN]=\sum_{\nd=1}^\Nd\vectornorm{\Unsn-\ebNn\Unsn}^2 + \vectornorm{\ebNn\Utsn}^2
  \label{equ:learn-n-with-tasks}
\end{equation}
as illustrated in \fref{fig:obj-uts-uns}. While this incurs a slight increase in computational cost (due to the need to evaluate the two terms instead of one), it has important benefits in ensuring the learnt $\ebA$ has \emph{correct rank}, that are missed if~\eref{equ:learn-n-uns} or~\eref{equ:learn-n-uts} are used in isolation. In other words \eref{equ:learn-n-with-tasks} helps ensuring the learnt constraints have the \emph{correct dimensionality}, as shall be illustrated in the following.

\begin{figure}[t!] 	
  \centering  
  \subfloat[] {\label{fig:minimise-uns}\includegraphics[width=4cm]{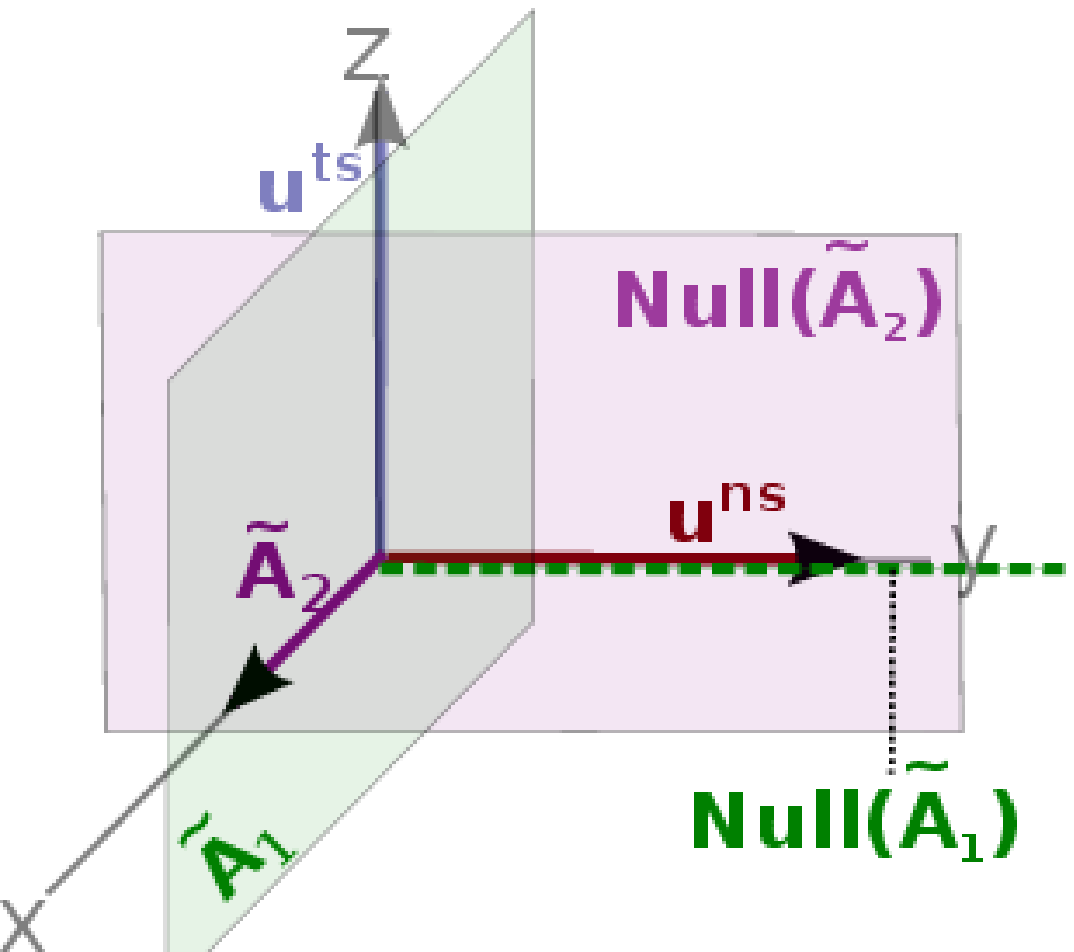}} 
  \subfloat[] {\label{fig:minimise-uts}\includegraphics[width=4cm]{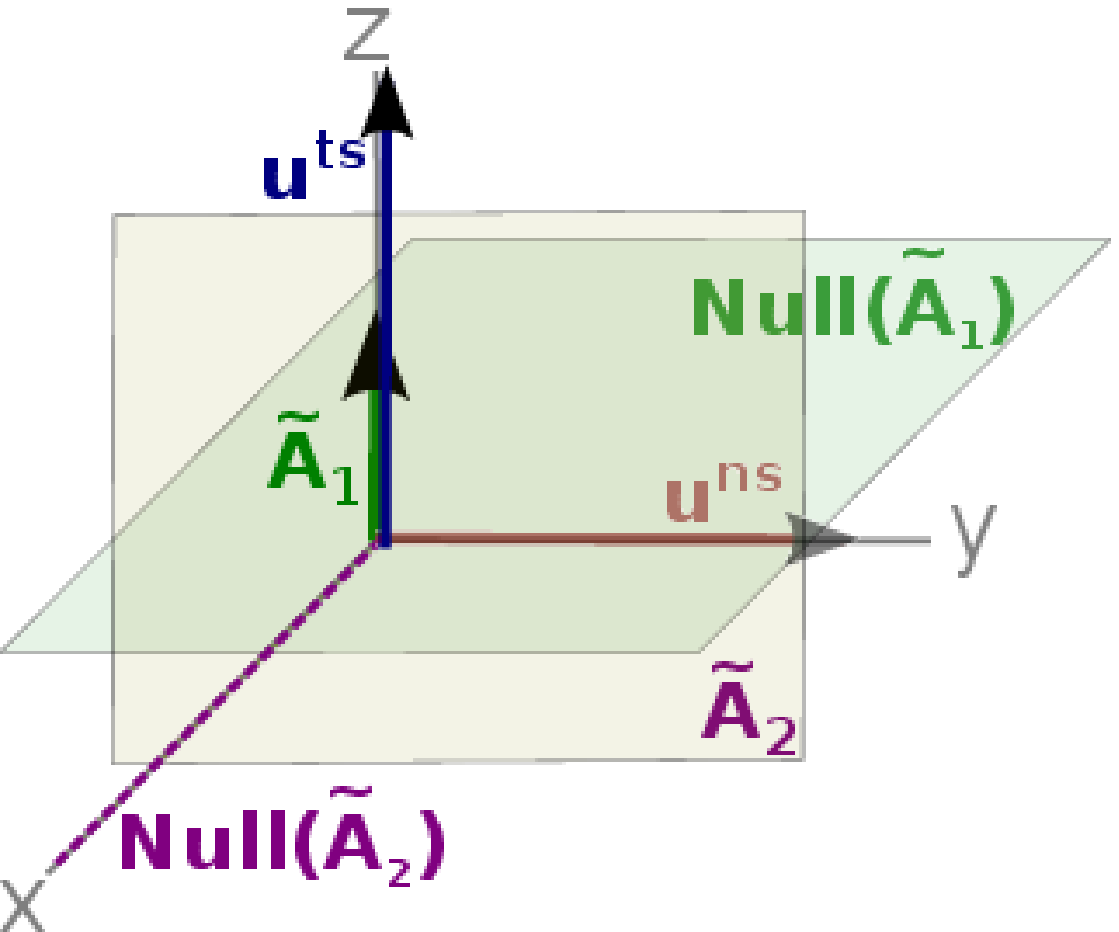}} \vspace{-3mm}
  \caption{Issues with learning $\ebN$ through~\eref{equ:learn-n-uns} or~\eref{equ:learn-n-uts}. (a) Using~\eref{equ:learn-n-uns}, both  $\ebA_1$ ($xz$-plane) and $\ebA_2$ ($x$-axis) are candidate solutions. Note that, $\ebA_2$ is incorrect since $\ebN_2\buts=\buts$, but this is not evident from the error measure.  (b) Using~\eref{equ:learn-n-uts} instead, $\ebA_1$ (on the $z$-axis) and $\ebA_2$ (the $yz$-plane) are candidate  solutions. However, $\ebA_2$ is wrong because $\ebN_2\buns=\bO$.  }
  \label{fig:minimise-uns-uts} 
\end{figure}
\subsection{Ensuring Correct Rank}
\noindent%
Consider again the example data point illustrated in \fref{fig:obj-data}, with $\buts$ (blue) parallel to the $z$-axis and $\buns$ (red) parallel to the $y$-axis. Minimising \eref{equ:learn-n-uns} or \eref{equ:learn-n-uts} in isolation for this data point has subtly different results.

\subsubsection{Minimising~\eref{equ:learn-n-uns}}

Recall that~\eref{equ:learn-n-uns} relies on finding $\ebN$ such that $\buns$ lies in the image space of $\ebN$, or
$\buns\in Image(\ebN)$. However, the problem is that the superset of $\bN$
also minimises this error measure. Let $\ebN'$ be a superset of $\ebN$ such
that $\ebN\subset\ebN'$. Since $\buns\in Image(\ebN)$ and $\ebN\subset\ebN'$,
then it is also true that $\buns\in Image(\ebN')$. In other words, the rank of $\ebN$
can be \emph{overestimated} by minimising \eref{equ:learn-n-uns}.

This is visualised in \fref{fig:minimise-uns}. There, two candidate solutions are (i) $\ebA_1$ ($xz$-plane), with the $y$-axis as the null space, and (ii) $\ebA_2$ ($x$-axis) with the $yz$-plane as null space. Note that both minimise \eref{equ:learn-n-uns} exactly, so without prior knowledge of the task-space nor dimensionality $\dimB$, it is hard to decide which one is the better estimate. If the data contains rich enough variations in $\buns$ such that $\buns$ spans $\Real^{\dimU-\dimB}$, then the $\ebN$ with the lowest rank, can heuristically be chosen, but this may be hard to verify in real world problems. Meanwhile, note that $\ebA_2$ is not a correct solution since $\ebN_2\buts=\buts$. 

\subsubsection{Minimising~\eref{equ:learn-n-uts}}
The converse problem arises by optimising~\eref{equ:learn-n-uts} instead. By definition, there exists a solution $\ebN$ in $\R^{\dimU-\dimB}$ such that $\ebN\buts=\bO$. However, note also that there also exist subspaces of $\ebN$ that satisfy this condition. Specifically, \eref{equ:learn-n-uts} seeks a projection matrix $\ebN$ such that its image space is orthogonal to $\buts$. 

The rank of the true solution $\bN$ is $\dimU-\dimB$, which implies that the image space of $\bN$ can be described by $\dimU-\dimB$ linearly independent row vectors. Note that, if we find a $\ebN$ such that $\ebN$  is orthogonal to $\buts$, then each row vector of $\ebN$, and their linear combinations, are also orthogonal to $\buts$. 

The issue is shown graphically in \fref{fig:minimise-uts}. There, candidate solutions $\ebA_1$ ($z$-axis) and $\ebA_2$ ($yz$-plane) both minimise~(\ref{equ:learn-n-uts}). However, $\ebA_2$ is wrong because $\ebN_2\buNs=\bO$.  The risk here is \emph{underestimating} the rank of $\ebN$. If it is known that $\buts$ spans the task-space, then the $\ebN$ with the highest rank (or $\ebA$ with the lowest rank) can be chosen, but this again relies on a heuristic choice.

\subsubsection{Minimising~\eref{equ:learn-n-with-tasks}}
If we have rich enough observations such that $\buts$ spans $\Real^{\dimB}$ and $\buns$ spans $\Real^{\dimU-\dimB}$, there is a projection matrix $\bN\in\Real^{(\dimU-\dimB)\times\dimU}$ such that
$\bN\buns=\buns$ and $\bN\buts=\bO$. From the prior analysis, if $\ebN\buns=\buns$ is satisfied, a projection matrix $\ebN$ such that $\bN\subseteq \ebN$ has been found. Likewise, if $\ebN\buts=\bO$, then it is also true that that $\ebN \subseteq\bN$. If both conditions are met, $\ebN\subseteq\bN\subseteq\ebN$, and the only possibility is $\ebN=\bN$. Therefore, \eref{equ:learn-n-with-tasks} can be applied to ensure that our estimated $\ebN$ has the current rank.

Assuming that $\bAn$ is formed from a set of $\dimB$ row vectors 
\[\bAn = [\ba_1(\bxn)^\T, \ba_2(\bxn)^\T,...,\ba_\dimR(\bxn)^\T]^\T\]
where $\ba_s$ corresponds to the $s^{th}$ constraint in the observation. $\ebA$ can be estimated by an iterative approach, whereby a series of $\ba_s$ are fitted to form an estimate $\ebA$~\cite{Lin.2015.ICRA}. Specifically, the $s^{th}$ constraint vector $\ba_s$ is learnt by optimising \eref{equ:learn-n-uns}, and $\ba_s$ is added only if it does not reduce the fit under \eref{equ:learn-n-with-tasks}. The process is summarised in Algorithm~\ref{alg:learn-n-task}. Note that, \emph{no prior knowledge of the true projection matrix $\bN$ is required}. 

\begin{algorithm}[t!]
    \caption{Learning Null-space Projection}
    \begin{algorithmic}[1]
    \REQUIRE \hfill \\
	$\bx$: observed states \\
	$\bu$: observed actions   
    \ENSURE \hfill \\
      $\ebA$: the estimated constraint matrix \\    
    \STATE Decompose $\bu$ into $\LearntUts$ and $\LearntUns$ by~(\ref{equ:learn-uns})
    \STATE Set $\ebA \leftarrow \emptyset$ and $\nb\leftarrow1$ 
    \STATE Learn $\ba_1^*$ by minimising~\eref{equ:learn-n-uns} 
    \WHILE{Adding $\ba_s^*$ to $\ebA$ does not increase \eref{equ:learn-n-with-tasks}}
      \STATE Set $\ebA\leftarrow[\ba_1^*,\cdots,\nba_{\nb}^*]^\T$ and $\nb\leftarrow\nb+1$         
      \STATE Learn $\ba_{\nb}^*$ by~\eref{equ:learn-n-uns} such that $\ba_{\nb} \perp \ba_j^*\,~\forall j<\nb$     
    \ENDWHILE
    \STATE Return $\ebA$
    \end{algorithmic}
    \label{alg:learn-n-task}
\end{algorithm}
  
\subsection{Learning constraints in operational space formulation} \label{sec:learn-vector}
\noindent
In the context of operational space formulation~\cite{1987.IJRR.Khatib}, a constraint can be imposed in the operational/task space , which has a one-to-one relationship with the state space. For example, given a constraint in the end-effector (task space), produce the joint movement (state space) that satisfied the constraints. We proposed two methods for learning the constraints in this formulation.

\subsubsection{Learning the selection matrix $\bLambda$}
\label{sec:search-lambda}
One way to represent $\bAn$ is to assume that $\bAn= \bLambda\Jacobian_n$ where $\bLambda$ is a selection matrix specifying which dimensions are constrained (i.e., $\bLambda_{s,s}=1$ if the $s^{th}$ dimension of the end-effector is constrained), and $\Jacobian_n$ is the Jacobian matrix that relates the joint velocity to the end-effector velocity. If $\Jacobian$ is known, we can form $\bLambda$ as a set of $\dimB$ orthonormal vectors $\bLambda = [\bLambda_1^\T, \bLambda_2^\T,...,\bLambda_\dimR^\T]^\T$ where $\bLambda_s \in \Real^{1\times \dimR}$ corresponds to the $s^{th}$ dimension in the task space and $\bLambda_i \perp \bLambda_j $ for $i \neq j$.

From~\cite{Lin.2015.ICRA}, the objective function in~(\ref{equ:learn-n-uns}) can be written as \[E[\ebN]=\sum_{n=1}^{\Nd}\ebunsn\Transpose\pinv{\ebAn}\ebAn\ebunsn.\] Substituting $\bAn = \bLambda\Jacobian_n$, \eref{equ:learn-n-uns} can be written as
\begin{equation}    
  E[\ebN] =\sum_{\nd=1}^\Nd (\ebunsn)^\T \pinv{(\bLambda\Jacobian_n)}(\bLambda \Jacobian_n) \ebunsn  
  \label{equ:learn-lambda}
\end{equation}
Then, the optimal $\bLambda$ can be formed by iteratively searching the choice of $\bLambda_s$ that minimises the \eref{equ:learn-lambda}.

Following~\cite{Lin.2015.ICRA}, an unit vector $\nba=(\na_1,\na_2,\cdots,\na_\dimU)$ with an arbitrary dimension $\dimU$ can be represented by $\dimU-1$ parameters $\btheta=(\theta_1,\theta_2,\cdots,\theta_{\dimU-1})^\T$ where
\begin{equation}
\begin{aligned}
  \na_1         &= \cos\theta_1,                          ~        		 
  \na_2         = \sin\theta_1\cos\theta_2,                ~    \\ 		 
  \na_3         &= \sin\theta_1\sin\theta_2\cos\theta_3 \dots	\\		 
  \na_{\dimU-1} &= \prod_{\nu=1}^{\dimU-2} \sin\theta_{\nu}\cos\theta_{\dimU-1},~ 
  \na_{\dimU}   = \prod_{\nu=1}^{\dimU-1} \sin\theta_{\nu}       
\end{aligned}
\label{equ:alpha-representation}
\end{equation}

\noindent Using the formulation above, each $\bLambda_s$ can be represented by parameters $\btheta_s \in \Real^{\dimU-1}$. Note that estimating $\btheta_s^*$ is a non-linear least squares problem, which cannot be solved in closed form. We use the Levenberg-Marquardt algorithm, a numerical optimization technique, to find the optimal $\btheta_s^*$.

\subsubsection{Learning state dependent constraint vector $\ba(\bxn)$}
\noindent In case Jacobian is not available, the second approach considers directly learning the constraint vector $\ba$. Assuming that $\bAn$ is formed from a set of $\dimB$ unit vectors $\bAn = [\nba_1(\bxn), \nba_2(\bxn),...,\nba_\dimR(\bxn)]^\T$ where $\nba_s$ corresponds to the $s^{th}$ constraint in the observation and $\nba_i \perp \nba_j $ for all $i \neq j$. 

\noindent Similar to the procedure in \sref{sec:search-lambda}, we can represent each $\nba_s$ as a set of $\dimU-1$ parameters $\theta_s$ as described in \eref{equ:alpha-representation}. For parameter estimation, $\bLambda_s$ is modeled as $\btheta_s (\bxn)= \WeightsMatrix_s \Basis(\bxn)$ where $\WeightsMatrix_s \in \Real^{(\dimR-1) \times \dimPhi}$ is a matrix of weights, and $\Basis(\bxn) \in \Real^{\dimPhi}$ is a vector of $\dimPhi$ fixed basis functions. We chose the normalised radial basis functions $\Basis_i(\bxn) = \frac{K(\bxn-c_i)} { \sum_{m=1}^{\dimPhi} K(\bxn-c_m)}$ where $K(.)$ denotes Gaussian kernels and $c_i$ are $\dimPhi$ pre-determined centres chosen according to k-means.

      

\section{Evaluation}        \label{evaluation}        \noindent In this section, some numerical results are presented to evaluate the
learning performance.

\subsection{Evaluation Criteria}\label{evaluation_criteria}\noindent
\noindent The goal of this work is to predict the projection matrix $\bN$ of the underlying the constrained observations so that these may be reproduced through a suitable learning scheme (\eg\cite{Howard.2009.AR,Towell.2010.IROS,Howard2009c}).  For testing the performance of learning, the following evaluation criteria may be defined.

\subsubsection{Normalised Projected Policy Error}
This error measure measures the difference between the policy subject to the
true constraints, and that of the policy subject to the estimated constraints.
Formally, the \emph{normalised projected policy error} (NPPE) can be defined as
\begin{equation}	
    E_{PPE} = \frac{1}{\Nd\sigma^2_{\bu}}\sum_{\nd=1}^\Nd \vectornorm{\bNn\bpin-\ebNn\bpin}^2
    \label{equ:NPPE}
\end{equation}
where $\Nd$ is the number of data points, $\bpin$ are samples of the policy,
and $\bN$ and $\ebN$ are the true and the learnt projection matrices,
respectively. The error is normalised by the variance of the observations $\sigma^2_\bu$.

\subsubsection{Normalised Projected Observation Error}
To evaluate the fit of $\ebN$ under the objective function~(\ref{equ:learn-n-with-tasks}), we suggest to use the \emph{normalised projected observation error} (NPOE),
\begin{equation}	
  E_{POE} = \frac{1}{\Nd\sigma^2_{\bu}}\sum_{\nd=1}^\Nd\vectornorm{\bunsn-\ebNn\bunsn}^2+\vectornorm{\ebNn\butsn}^2
  \label{equ:NPOE}	
\end{equation}
which reaches zero only if the model exactly satisfies \eref{equ:learn-n-with-tasks}.

\subsubsection{Normalised Null Space Component Error}
\label{ss:nnce}
Since the quality of the input data to our proposed method depends on the fitness of $\ebuns$, we also suggest looking at the \emph{normalised null space component error} (NNCE), 
\begin{equation}
  E_{NCE} = \frac{1}{\Nd\sigma^2_{\bu}}\sum_{\nd=1}^\Nd\vectornorm{\bunsn-\ebunsn}^2
  \label{equ:NNCE}
\end{equation}
which measures the distance between the true and the learnt null space components $\buns$ and $\ebuns$, respectively.

\subsection{Toy Example}
\noindent Our first experiment demonstrates our approach on a two-dimensional system with a one-dimensional constraint ($\bA\in\R^{1\times2}$). We consider three different null space policies $\bpi$:
\il{\medmuskip=0mu
  \item a linear policy: $\bpi=-\bL\bar{\bx}$ where $\thinmuskip=0mu\bar{\bx}:=(\bx^\T,1)^\T$ and $\!\bL=\left((2,4,0), (1,3,-1)\right)^\T$, 
  \item a limit-cycle policy: $\dot{r}=r(\rho-r^2)$ with radius $\rho\,=\,0.75\,m$, angular velocity $\dot{\theta}\,=\,1\,rad/s$, where $r$ and $\phi$ are the polar representation of the state, \ie $\bx=(r\cos\phi, r\sin\phi)^\T$, and 
  \item a sinusoidal policy: $\bpi=(\cos{z_1}\cos{z_2}, -\sin{z_1}\sin{z_2})^\T$ with $z_1=\pi x_1$ and $z_2=\pi(x_2+\frac{1}{2})$.
}

The training data consists of $150$ data points, drawn randomly across the space $(\bx)_i\sim\U(-1,1),\,i\in\{1,2\}$. The data points are subjected to a $1$-D constraint $\bA=\nba\in\R^{1\times2}$, in the direction of the unit vector $\nba$, which is drawn uniform-randomly $\theta\sim\U(0,\pi]\,rad$ at the start of each experiment.

\begin{figure}[t!]%
  \centering%
  \includegraphics[width=.5\linewidth]{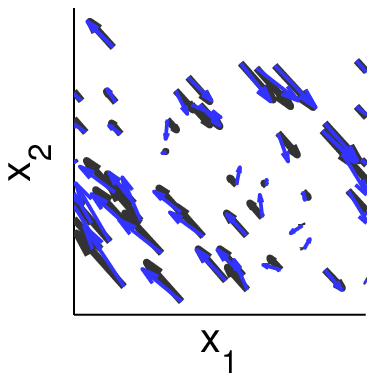}\mylabel{fig:toy-train-data}{(left)}~%
  \includegraphics[width=.5\linewidth]{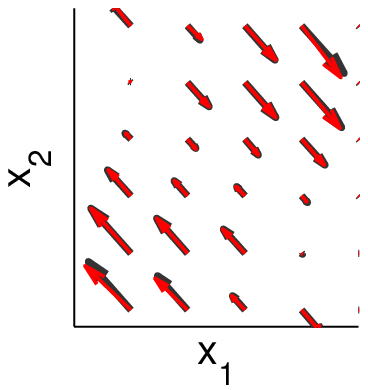}\mylabel{fig:toy-vis-data}{(right)}\vspace{-2mm}
  \caption{Samples of training data generated from a limit-cycle policy \ref{fig:toy-train-data} and a grid of test points \ref{fig:toy-vis-data}. Shown are the true null space component $\buns$ (black), the predicted null space component $\ebuns$ (red), and the true observation $\bu$ (blue).}
  \label{fig:visualise-toy}  
\end{figure}

The underlying task is to move with fixed velocity to the target point $\rho^*$ along the direction given by $\nba$. This is achieved with a linear attractor policy in task space
\begin{equation}
  \bb(\bx) = \beta^{ts} (\rho^* - \rho)
  \label{equ:task-space-policy}
\end{equation}
where $\rho$ denotes the position in task space, and $\beta^{ts} = 0.1$ is a scaling factor. For each trajectory, the task space target was drawn randomly $\rho^* \sim \U[-2,2]$. A sample data set for the limit-cycle policy is presented in Fig.~\ref{fig:visualise-toy} (left) where the colours denote the true null space component $\buns$ (black) and the true observation $\bu$ (blue).

The null space component is modeled as $\LearntUns = \WeightsVector \Basis$ where $\Basis$ is a vector of $\dimPhi = 16$ radial basis functions, and $\WeightsVector$ is a vector of weights learnt through minimisation of the objective function~\eref{equ:learn-uns}. Then, the projection matrix $\ebN$ is learnt by minimising~(\ref{equ:learn-n-uns}), according to the scheme outlined in \S\ref{method}. The experiment is repeated 50 times and evaluated on a set of 150 test data points, generated through the same procedure as described above. 

\begin{table}[t!]
	\centering
	\begin{tabular}{*{4}{l}}      
		\hline
		Policy      & NNCE                         & NPPE                         & NPOE                         \\ 
		\hline 
		Linear	    & $\sim10^{-7}$ & $\sim10^{-9}$& $\sim10^{-9}$\\ 
		Limit-cycle & $0.08\pm0.02$ & $0.001\pm0.002$ & $0.001\pm0.002$ \\ 
		Sinusoidal  & $5.26\pm4.46$ & $0.011\pm0.017$ & $0.014\pm0.021$ \\ 
		\hline 
	\end{tabular}
	\caption{NNCE, NPPE, and NPOE (mean$\pm$s.d.)$\times10^{-2}$ over 50 trials.}
	\label{table:experiment-toy}\vspace{-1ex}
\end{table}

\tref{table:experiment-toy} summarises NPPE, NPOE, and NNCE (\eref{equ:NPPE}-\eref{equ:NNCE}) for each policy. The values are (mean$\pm$s.d.) over $50$ trials on the hold-out testing set. In terms of NPPE and NPOE, we can learn a good approximation with both measurement $< 10^{-4}$ without the true decomposition of $\buTs$ and $\buNs$. Looking at the result of NNCE, we note that quality of $\ebN$ is affected by the accuracy of $\ebuns$. However, if the true $\buns$ are given, NPPE and NPOE are both lower than $10^{-15}$, and this is not significantly affected by the policy. 

The results for the limit-cycle policy are shown in \fref{fig:visualise-toy}~\ref{fig:toy-vis-data}. The predicted $\ebuns$ (red) are plotted on top of the true $\buns$ (black). As can be seen, there is good agreement between the two, verifying that using $\ebuns$, there is little degradation in predicting constrained motion.

\begin{figure}[t!]%
	\centering%
	\includegraphics[width=.5\linewidth]{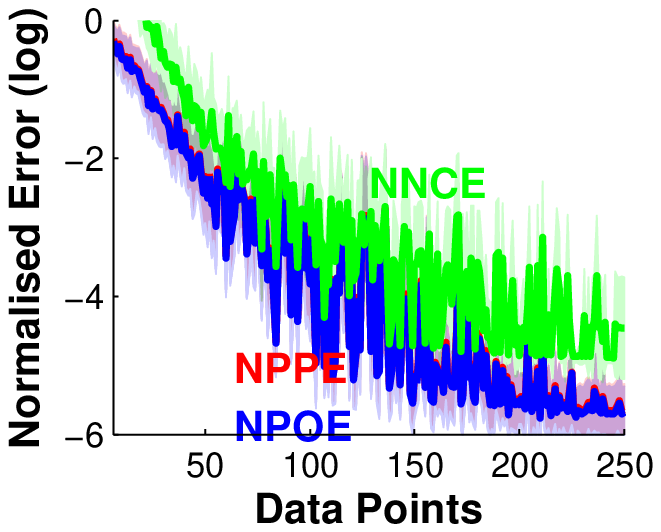}\mylabel{fig:experiment-toy-point}{(left)}~%
	\includegraphics[width=.5\linewidth]{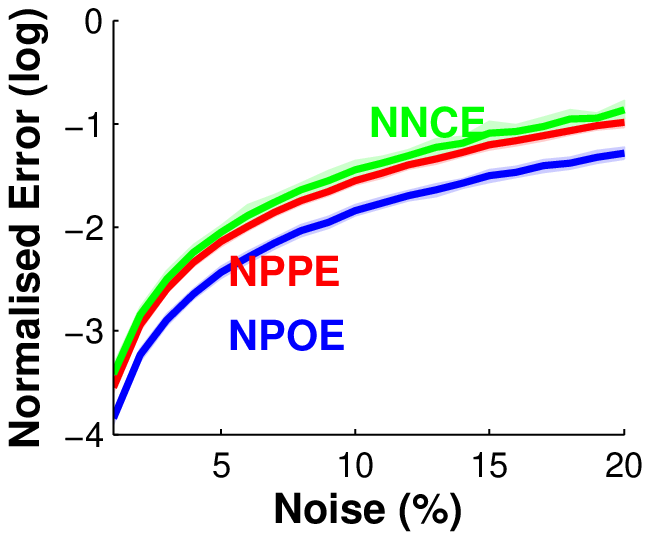}\mylabel{fig:experiment-toy-noise}{(right)}\vspace{-2mm}
    \caption{NNCE, NPPE, and NPOE for increasing number of data points \ref{fig:experiment-toy-point} and increasing noise levels in the observed $\action$ \ref{fig:experiment-toy-noise}. Curves are mean$\pm$s.d. over 50 trials.}
    \label{fig:experiment-toy-PPE-POE}
\end{figure}

To further characterise the performance of the proposed approach, we also looked at the effect of varying the size of the input data for the limit-cycle policy. We tested our method for $5<\Nd<250$ data points and estimated $\LearntN$ from the learnt $\buTs$ and $\buNs$. The results (in log scale) over 50 trials are plotted in \fref{fig:experiment-toy-PPE-POE}~\ref{fig:experiment-toy-point}. It can be seen that the NPPE, NPOE, and NNCE rapidly decrease as the number of input data increases. This is to be expected, since a data set with richer variations can form a more accurate estimate of $\LearntUts$, $\LearntUns$, and $\LearntN$. Note that even at relatively small data set ($\Nd<50$), the error
is still very low ($\approx 10^{-3}$). 

We also tested how the levels of noise in the training data affect the
performance of our method. For this, we contaminated the limit-cycle policy
$\bpi$ with Gaussian noise, the scale of which we varied to match up to $20\%$
of the data. The resulting NNCE, NPPE, and NPOE follows the noise level, as
plotted in \fref{fig:experiment-toy-PPE-POE}~\ref{fig:experiment-toy-noise}. It
should be noted, however, that the error is still relatively low (NPPE
$<10^{-2}$), even when the noise is as high as $5\%$ of the variance of the
data.

\subsection{Three Link Planar Arm}\label{label:experiment-3link}
\noindent The goal of the second experiment is to assess the performance of the proposed approach for more realistic constraints. For this, constrained motion data from a kinematic simulation of a planar three link arm is used.

The set up is as follows. The state and action spaces of the arm are described by the joint angles $\bx:=\bq\in\Real^3$ and the joint velocities $\bu:=\bqdot\in\Real^3$. The task space is described by the end-effector $\br=(\r_x,\r_z,\r_\theta)^\T$ where $\r_x$ and $\r_z$ specify the position, and $\r_\theta$ is the orientation.

Joint space motion is recorded as the arm performs tasks under different constraints in the end-effector space. As discussed in \sref{sec:search-lambda}, a task constraint $\bA$ at state $\bxn$ is described through the form 
\begin{equation}
    \bAn =\bLambda\bJ_n
\end{equation}
where $\bJ_n\in\R^{3\times3}$ is the manipulator Jacobian, and $\bLambda\in\R^{3\times3}$ is the selection matrix specifying the coordinates to be constrained. The following three different constraints can be defined:
\begin{enumerate}
\item $\bLambda_{(x,z)}     =((1,0,0),(0,1,0),(0,0,0))^\T$,
\item $\bLambda_{(x,\theta)}=((1,0,0),(0,0,0),(0,0,1))^\T$, and
\item $\bLambda_{(z,\theta)}=((0,0,0),(0,1,0),(0,0,1))^\T$.
\end{enumerate}
Choosing $\bLambda_{(x,z)}$ allows the orientation to follow the null space policy $\bpi$ provided that the end-effector position moves as required by the task space policy $\TaskspacePolicy$. Constraint $\bLambda_{(x,\theta)}$ restricts the space defined by the x-coordinate $\r_x$ and orientation of the end-effector $\r_\theta$, while $\r_z$ is unconstrained.
        
\begin{figure}[t!]
    \centering  
    \includegraphics[width=.33\linewidth]{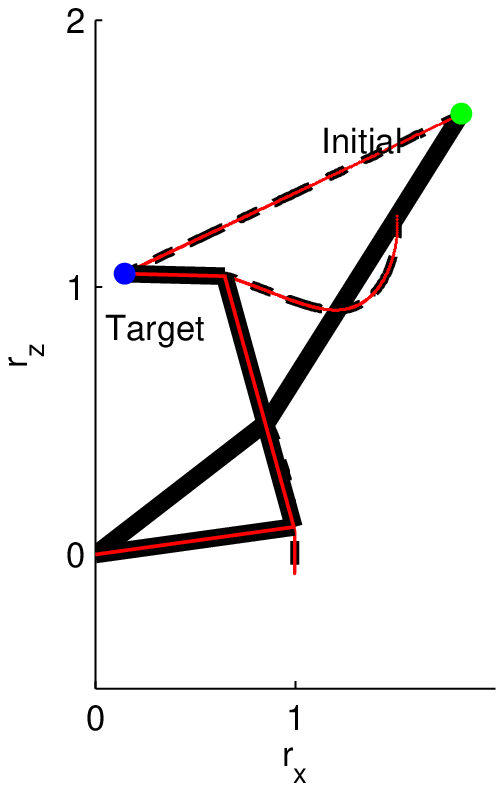}\mylabel{fig:3link-t}{(left)}~%
    \includegraphics[width=.33\linewidth]{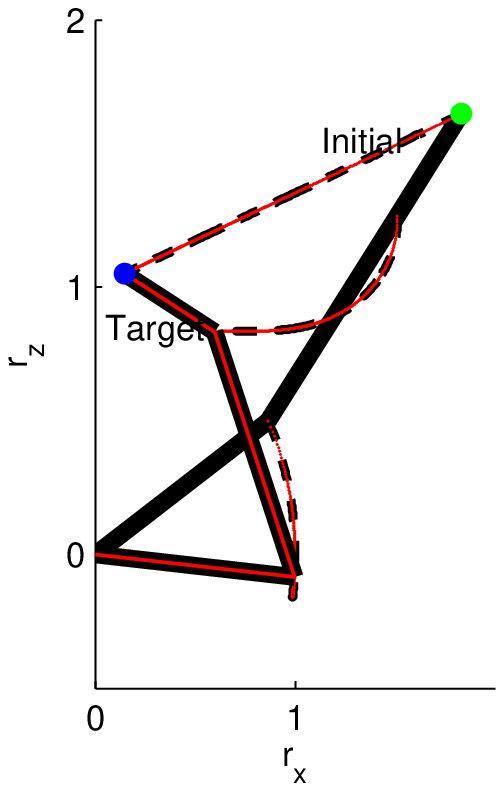}\mylabel{fig:3link-p}{(centre)}~%
    \includegraphics[width=.33\linewidth]{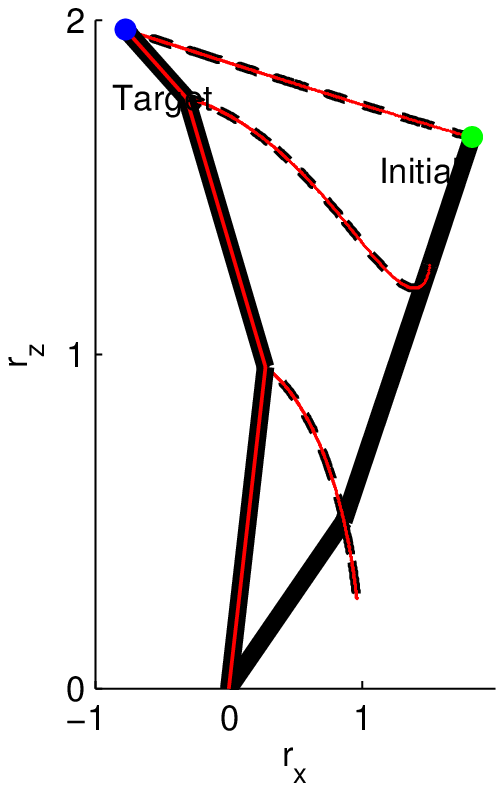}\mylabel{fig:3link-b}{(right)}\vspace{-2mm}
	\caption{Example trajectories generated from testing data \ref{fig:3link-t}, a new null space policy \ref{fig:3link-p}, and a new task space policy \ref{fig:3link-b}. Shown are trajectories using the true $\bA$ (black) and the learnt $\ebA$ (red).}
    \label{fig:3link-traj}
\end{figure}

The trajectories are recorded from a null space policy $\bpi=-\bL(\bx-\bx^*)$ where $\bx^*=\bO$ and $\bL=\I$ under the active constraint. In each trajectory, the task space policy is a linear policy tracking a task space target $\br^*$, which is drawn uniformly from $\br_x^* \sim U[-1,1]$, $\br_z^* \sim U[0,2]$, $\br_\theta^* \sim U[0,\pi]$. The start states for each trajectory were randomly selected from the uniform distribution $\bq_1 \sim U [0\Degree,10\Degree], \bq_2 \sim U[90\Degree,100\Degree],\bq_3 \sim U[0\Degree,10\Degree]$. Targets without a valid inverse kinematics solution are discarded. For each task constraint, 50 trajectories each of length 50 steps were generated. Following the same procedure, another 50 trajectories are also generated as unseen testing data. \fref{fig:3link-traj}~\ref{fig:3link-t} shows an example trajectory with constraint $\bLambda_{x,z}$ and task space target $\br^*=[1,0]$ (black line).

In a real situation, the null space component is $\LearntUns$ is unlikely to be available in the raw data. Therefore, a parametric model of the form $\LearntUns = \WeightsVector \Basis$ is learnt from this data by minimising \eref{equ:learn-uns}. Here, $\Basis$ is a vector of $100$ radial basis functions, and $\WeightsVector$ is a vector of parameters. The latter is then used to learn the constraint $\bA$ through the methods outlined in~\sref{sec:learn-vector}. In the following results for 50 repeats of this experiment are reported.

\begin{table}[t]
	\centering%
	\begin{tabular}{*{5}{c}}
		\hline
		Constraint            & NNCE          & Method           & NPPE & NPOE \\ \hline		     
		$\bLambda_{x,z}$      & $0.32\pm1.07$ & Learn $\nba$     & $3.54\pm2.29$ & $0.06\pm0.04$ \\ 
		                      &               & Learn $\bLambda$ & $0.30\pm0.44$ & $0.04\pm0.04$ \\ 
		\hline                                     
		$\bLambda_{x,\theta}$ & $3.89\pm1.10$ & Learn $\nba$     & $7.79\pm12.0$ & $2.14\pm6.48$ \\ 
		                      &               & Learn $\bLambda$ & $0.32\pm1.42$ & $2.53\pm1.11$ \\ 
		\hline		                          
		$\bLambda_{z,\theta}$ & $0.61\pm0.60$ & Learn $\nba$     & $2.80\pm2.59$ & $0.43\pm0.24$ \\ 
		                      &               & Learn $\bLambda$ & $0.24\pm0.51$ & $0.14\pm0.25$ \\ 
		\hline
	\end{tabular}
	\caption{NNCE, NPPE, and NPOE (mean$\pm$ s.d.)$\times10^{-2}$ for each constraint over 50 trials.}
	\label{table:experiment-3link}
\end{table}

Looking at the NPPE and NPOE columns in \tref{table:experiment-3link}, a good approximation of the null space projector is learnt for each of the constraints, with errors of order $10^{-2}$ or lower in all cases. It can also be seen that the errors when learning $\bLambda$ are lower than those when learning $\nba$. This is to be expected since the former relies on prior knowledge of $\Jacobian(\state_n)$, while the latter models the nonlinear, state dependent $\bLambda\Jacobian(\state_n)$ in absence of this information. 

To further test the accuracy of the approximation, the trajectory generated under the learnt constraints can be compared with those under the ground truth constraints,  using the same task and null space policies. In \fref{fig:3link-traj}~\ref{fig:3link-t}, the red line shows this for the aforementioned example trajectory with learnt $\tilde{\bLambda}_{x,z}$. As can be seen, the latter matches the true trajectories extremely well. 

Finally, to test the ability of the learnt approximation to generalise to new situations, the trajectories generated with (previously unseen) %
(i) null space policy $\bpi'$ and
(ii) tasks $\bb'$ 
are also compared to see if the learnt constraint can be used to predict behavioural outcomes for policies not present in the training data. 

\fref{fig:3link-traj}~\ref{fig:3link-p} shows the trajectory generated when a new null space policy $\bpi'= -0.05 \vectornorm{\bx}^2$, not present in the training data, is subjected to (i) the true constraints (\ie $\pinv{\bA}\bb+\bN\bpi'$, black), and (ii) the learnt constraint (\ie $\pinv{\ebA}\bb+\ebN\bpi'$, red). \fref{fig:3link-traj}~\ref{fig:3link-b} shows the trajectory generated when a new task space policy $\bb'$ with a new task space target $\br'=[-1,2]$ under (i) the true constraints (\ie $\pinv{\bA}\bb'+\bN\bpi$, black), and (ii) the learnt constraint (\ie $\pinv{\ebA}\bb'+\ebN\bpi$, red). In both cases, a close match is seen between the predicted behaviour under the true and learnt constraints, indicating good generalisation performance.



\hspace{-5mm}
  
\section{Conclusion}        \label{conclusion}        \noindent In this paper, we consider how the null space projection matrix of a kinematically constrained system, and have developed a method by which that matrix can be approximated in the absence of any prior knowledge either on the underlying movement policy, or the geometry or dimensionality of the constraints.

Our evaluations have demonstrated the effectiveness of the proposed approach on problems of differing dimensionality, and with different degrees of non-linearity. The have also validated the use of our method in the adaptation of a learnt policy onto a novel constraints.

For future research, we plan to validate the proposed method on robots with higher degree of freedom and human data where the true policy and constraint are both unknown, and to study variants of the approach that may improve its efficiency through iterative learning approaches.

%
\bibliographystyle{IEEEtran}
\bibliography{abbreviations,paper}
\end{document}